\documentclass[sigconf]{acmart}
\renewcommand\footnotetextcopyrightpermission[1]{}    
\AtBeginDocument{%
  }

\usepackage{amsmath}
\usepackage{graphicx} 
\usepackage{microtype}
\usepackage{booktabs} 
\usepackage{multirow}
\usepackage{graphicx} 
\usepackage{float}
\usepackage{bookmark}
\usepackage{amsfonts}
\usepackage{amsmath}
\usepackage{adjustbox}
\usepackage[ruled,linesnumbered]{algorithm2e}
\copyrightyear{2022}
\acmYear{2022}
\setcopyright{rightsretained}
\acmConference[KDDCup '22]{To be presented at KDD Cup 2022 Workshop:
ESCI Challenge for Improving Product Search}{August 17, 2022}{Washington, DC, USA}
\acmBooktitle{KDD Cup 2022 Workshop: ESCI Challenge for Improving Product Search, August 17, 2022, Washington, DC, USA}
\acmDOI{}
\acmISBN{}
\begin{document}

\title{ZhichunRoad at Amazon KDD Cup 2022: MultiTask Pre-Training for E-Commerce Product Search}

\author{Xuange Cui}
\email{cuixuange@jd.com}
\affiliation{%
  \institution{JD.com}
  \city{Beijing}
  \country{China}
}

\author{Wei Xiong}
\email{xiongwei9@jd.com}
\affiliation{%
  \institution{JD.com}
  \city{Beijing}
  \country{China}
}

\author{Songlin Wang}
\email{wangsonglin3@jd.com}
\affiliation{%
  \institution{JD.com}
  \city{Beijing}
  \country{China}
}


\begin{abstract}
In this paper, we propose a robust multilingual model to improve the quality of search results.
Our model not only leverage the processed class-balanced dataset, but also benefit from multitask pre-training that leads to more general representations.
In pre-training stage, we adopt mlm task, classification task and contrastive learning task to achieve considerably performance.
In fine-tuning stage, we use confident learning, exponential moving average method (EMA), adversarial training (FGM) and regularized dropout strategy (R-Drop) to improve the model's generalization and robustness.
Moreover, we use a multi-granular semantic unit to discover the queries and products textual metadata for enhancing the representation of the model.
Our approach obtained competitive results and ranked top-8 in three tasks.
We release the source code and pre-trained models associated with this work\footnote{\url{https://github.com/cuixuage/KDDCup2022-ESCI}}.
\end{abstract}



\maketitle

\section{Introduction}
With the rapid growth of e-Commerce, online product search has emerged as a popular and effective paradigm for customers to find desired products and engage in online shopping \cite{DBLP:journals/corr/abs-2001-04980, DBLP:journals/corr/abs-2106-09297, Liu2021Que2SearchFA}.
It is very challenging to accurately find and display relevant products.
This is because the customer queries are ambiguous and implicit \cite{DBLP:journals/corr/abs-2105-02978}.
For example, many users search for "iPhone" to find and purchase an "iPhone charger".
However, the traditional binary classification model is difficult to clearly characterize this relationship.
The Amazon KDD Cup 2022 presents a novel multilingual dataset \cite{reddy2022shopping} across English, Japanese and Spanish, and consists of three different subtasks to evaluate the model's abilities of ranking and classifying.

In this paper, our contributions can be summarized as follows:
1) Data Augmentation. We use the translation model to convert Spanish to English for expanding the dataset.
Through splitting the complement and irrelevant product text information, we could get a bigger dataset with balanced labels.
We use confident learning \cite{northcutt2017rankpruning, northcutt2021confidentlearning} to find the potential label errors and remove $\sim$4\% data from the training dataset.
2) MultiTask Pre-training. In pre-training stage, we use MLM task, classification task and contrastive learning task for improving the model's performance.
3) In fine-tuning stage, we use a multi-granular semantic unit to discover the queries and products textual metadata for enhancing the representation of the model.
And we observe that exponential moving average method(EMA) \cite{6708545}, adversarial training(FGM) \cite{goodfellow2015explaining} and regularized dropout strategy(R-Drop) \cite{DBLP:journals/corr/abs-2106-14448} could improve the model's generalization and robustness.

Our team participated in all tasks, and achieved considerably performance gain over the baseline solution.
Specifically, our approach ranked 5th in task1, ranked 7th in task2 and ranked 8th in task3.

\section{Background}
The Amazon KDD Cup 2022 \cite{reddy2022shopping} provides three subtasks.
The task1 consists of a query-product ranking task aimed at ranking the results list.
The Normalized Discounted Cumulative Gain(nDCG) \cite{DBLP:journals/corr/abs-1304-6480} will be used to evaluate the model's abilities of ranking.

The task2 and task3 are classification tasks which require the model to classify the query/product pairs into correct categories.
These tasks are designed to test the model's ability of classifying.
The micro-F1 \cite{DBLP:journals/corr/abs-1911-03347} will be used as an evaluation metric.
Moreover, the task2 consists of a multi-class product classification task aimed at classifying each product as being an Exact, Substitute, Complement, or Irrelevant match for the query.
The task3 will measure the model's abilities of identifying the substitute products in the list of results for a given query.  

The statistics of the corpus are shown in Table~\ref{tab:tabel_1}.
In this challenge, the organizers provide two different versions of the data set.
One for task 1 which is reduced version in terms of number of examples and ones for tasks 2 and 3 which is a larger \cite{reddy2022shopping}.
It is noted that the reduced version of the data set has more difficult samples.
Our team participated in all subtasks, and the next section will introduce an overview of our system.

\begin{table}
\centering 
\begin{tabular}{ccc|r}
    \toprule
    \textbf{SubTask} & \textbf{Train Dataset} & \textbf{Test dataset} & \textbf{Languages} \\
    \midrule
    Task1 & 781K & 48K & Spanish \\
    Task2 & 1834K & 277K & \& English \\
    Task3 & 1834K & 277K & \& Japanese \\
    \bottomrule
\end{tabular}
\caption{The statistics of datasets.}
\label{tab:tabel_1} 
\end{table}

\section{System Overview}
\subsection{Multi-Task Pre-Training}
We compare several pre-trained multilingual language models from the XTREME Leaderboard\footnote{\url{https://sites.research.google/xtreme}},
and then we use the "microsoft/infoxlm-large\footnote{\url{https://huggingface.co/microsoft/infoxlm-large}}" as text encoder.

\begin{figure}[htbp]
  \centering 
  \includegraphics[width=0.47\textwidth]{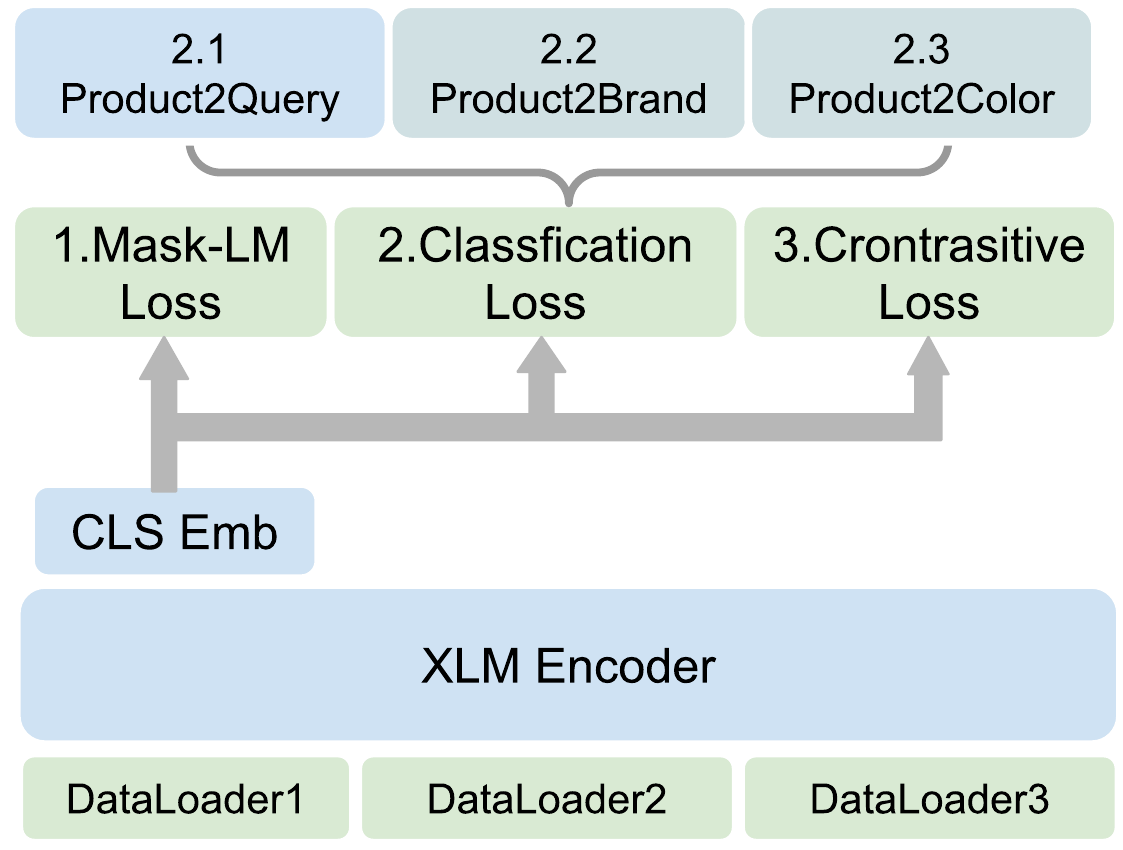}
  \caption{A schematic overview of our novel pre-training tasks. These tasks encourage the encoded representations to be more general.}
  \label{fig:figure_1}
\end{figure}

The InfoXLM$_{large}$ model \cite{DBLP:journals/corr/abs-2007-07834} containing 94 languages and pre-trained with CCNet dataset, and has the same configurations of XLM-R \cite{DBLP:journals/corr/abs-1911-02116} and a shared vocabulary size of 250002.
Figure~\ref{fig:figure_1} shows a high-level overview of our proposed pretext tasks.

\textbf{MLM Task}, is widely used for learning text representations \cite{DBLP:journals/corr/abs-1810-04805}. 
MLM trains a model to predict a random sample of input tokens that have been replaced by a [MASK] placeholder in a multi-class setting over the entire vocabulary \cite{DBLP:journals/corr/abs-2109-01819}.
We adopt MLM-Task on the multilingual product-catalogue dataset.

\textbf{Classification Task}, contains three classification subtasks. 
One of them is Product2Query-Task, this task is a binary classification task.
Based on the Poisson distribution, a piece of text is intercepted from commodity text information as the faked query.
The Parameters passed to the Poisson distribution and more details can be found in appendix \ref{appendix:Poisson}.
Product2Brand-Task and Product2Color-Task are multi-class classification that using product text information to predict the brand and the color of current item. 

\textbf{Contrastive Learning Task}, is mainly inspired by SimCSE \cite{gao2021simcse} and EsimCSE \cite{DBLP:journals/corr/abs-2109-04380}.
During training, each data point is trained to
find out its counterpart among $(N - 1)$ from in-batch negative samples and the queue of data samples.
The samples in the queue are progressively replaced.
\begin{equation}
  -\log \frac{e^{\operatorname{sim}\left(\mathbf{h}_{i}, \mathbf{h}_{i}^{+}\right) / \tau}}{\sum_{j=1}^{N} e^{\operatorname{sim}\left(\mathbf{h}_{i}, \mathbf{h}_{j}^{+}\right) / \tau}+\sum_{q=1}^{Q} e^{\operatorname{sim}\left(\mathbf{h}_{i}, \mathbf{h}_{q}^{+}\right) / \tau}}
\end{equation}
The $h_{*}$ is the sentence representation, where $h_{i}$ and $h_{i}^{+}$ are semantically related.
The $h_{q}^{+}$ denotes a sentence embedding in the momentum-updated queue.
And the $Q$ is the size of the queue, $sim(h1, h2)$ is the cosine similarity scores of sentence representations, $\tau$ is a temperature hyperparameter.
In the end, we average the all N Li losses to calculate the contrastive loss $\mathcal{L}_{\text{con}}$.
\begin{algorithm}
  \caption{Training a MultiTask model.}
  \label{algorithm:algorithm_1}
  \KwIn {DataSet $\mathcal{D}=\left\{\left(x, y, z\right)_{i}\right\}_{i=1}^{|\mathcal{D}|}$}
  Initialize model parameters $\Theta$ randomly \;
  Model trainer $T$ that takes batches of training data as input to  optimize the model parameters $\Theta$ \;
  Set the max number of epoch: $epoch_{\max}$ \;
  \For{$\text{epoch in }1, 2, ..., epoch_{\max}$}
  {Shuffle $\mathcal{D}$ by mixing data from different tasks \;
  \For{$\mathcal{B}\text{ in }\mathcal{D}$}{
    // $\mathcal{B}$ is a mini-batch of pre-training task \;
    Compute loss : $L(\Theta)$ \;
    1. $L(\Theta)$ = Mask LM Loss \;
    2. $L(\Theta)$ += Classification Loss \;
    3. $L(\Theta)$ += Contrastive Learning Loss \;
    Optimize the model using $L(\Theta)$ \;
  }
  }
  \KwOut {Pre-trained Model $\Theta$}
\end{algorithm}

\subsection{Fine-Tuning Methods}
\begin{figure*}[htbp]
  \centering 
  \includegraphics[width=0.85\textwidth]{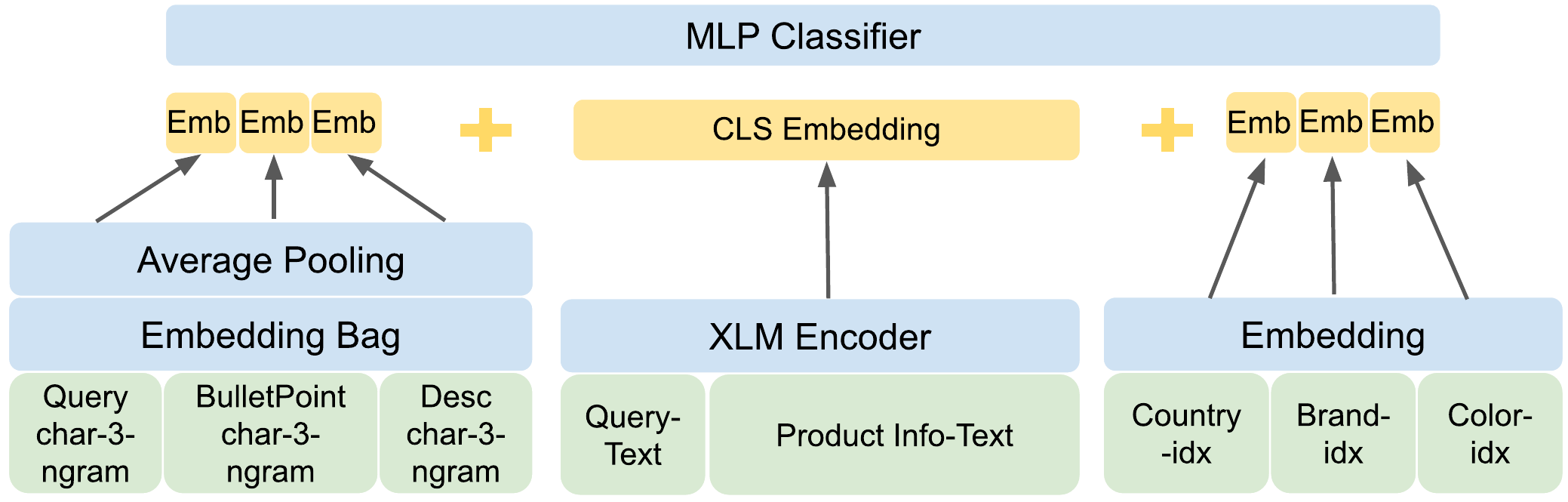}
  \caption{In fine-tuning stage, we concatenate the multi-granular semantic units, the [CLS] embedding from XLM encoder and the IDs' embeddings.}
  \label{fig:figure_2}
\end{figure*}

After pre-training, we remove the classifiers for pre-training multitask and concatenate some embeddings with an extra MLP classifier.
The embeddings consist of three sets of representations.
One of them is done by concatenating the queries' 3-gram mean-pooling, bullet points' 3-gram mean-pooling and descriptions' 3-gram mean-pooling embeddings.
The others consist of country embedding, brand embedding and color embedding, as shown in Figure~\ref{fig:figure_2}.

\textbf{Exponential Moving Average} Our model uses EMA \cite{6708545} to smooth the trained parameters.
Evaluations that use averaged parameters sometimes produce significantly better results than the final trained values.
Formally, we define the smoothed variables and trained variables as $\theta_{s}$ and $\theta_{t}$, EMA decay weight as: $\eta$.
After each training step, we update $\theta_{s}$ by:
\begin{equation}
    \theta_{s} \leftarrow \eta \theta_{s}+(1-\eta) \theta_{t}
 \end{equation}

 \textbf{Adversarial Training} Recently, adversarial attack has been widely applied in computer vision and natural language processing \cite{goodfellow2015explaining, madry2019deep, Zhu2020FreeLB, jiang-etal-2020-smart}.
 Many works use it during fine-tuning, we explore the influence of adversarial training strategies and compare the FGSM,
 PGD, FREELB and SMART methods.
 The adversarial attack works by augmenting the input with a small perturbation that maximizes the adversarial loss:
\begin{equation}
    \min _{\theta} \mathbb{E}_{(x, y) \sim \mathcal{D}}\left[\max _{\Delta x \in \Omega} L(x+\Delta x, y ; \theta)\right]
\end{equation}
where the $\mathcal{D}$ is dataset, $x$ is input, $y$ is the gold label, $\theta$ is the model parameters, $L(x, y; \theta)$ is the loss function and $\Delta x$ is the perturbation.
In our experiments, we adopt FGSM method in all tasks which based on the actual performances.

\textbf{R-Drop} is proved to be an effective regularization method based on dropout, by minimizing the KL-divergence of the output distributions of every two sub-models generated via dropout in model training.
\begin{equation}
  \mathcal{L}_{K L} = \alpha \cdot \left[ \mathcal{D}_{K L}\left(Logit_{1}, Logit_{2}\right) + \mathcal{D}_{K L}\left(Logit_{2}, Logit_{1}\right) \right]
\end{equation}
We use the origin logits of model's output as $Logit_{1}$, and the logits after adversarial attack as $Logit_{2}$.

\textbf{Embedding Mixup} is widely used data augmentation method through linearly interpolating inputs and modeling targets of random samples.
We use the contextual embedding vector of [CLS] and the corresponding label to generate synthetic examples for training.
Such training has been shown to act as an effective model regularization strategy for text classification task.
In conclusion, we present the self-supervised multitask pre-training tasks and the several fine-tuning methods for improving the models' generalization and robustness.

\section{Experiments}
\subsection{Settings}
We use InfoXLM$_{large}$ as the text encoder, the EMA decay weight is set to 0.999. 
And our learning rate is set to 1e-5 with warm-up ratio over 10\%, batch size is 32 and gradient clip norm threshold is set to 1. 
In pre-training stage, the maximum number of epochs was set to 10. And in the fine-tuning stage, the maximum number of epochs was set to 5. 
During adversarial training, we set $\varepsilon$  to 1.0 in FGM that means calculate only one step in the adversarial attack. 
We find that the dataset has imbalanced label and use some data processing steps. 
Through splitting the complement and irrelevant product text information, we could get more pairs which have the same label and get a more balanced dataset. 
We use confident learning to find the potential label errors and remove $\sim$4\% data from the training dataset. 
As presented in appendix \ref{appendix:Poisson}, the median of Spanish and English queries is 14 which satisfies the Poisson distribution of $\mu$ is set to 4.
And the median of the Japanese query is 31 which satisfies the Poisson distribution with $\mu$ is set to 8. 

\subsection{Main Results}
Our approach achieved considerably performance gain over the baseline solution, and ranked top-8 in three tasks.
The main results are shown in Table~\ref{tab:tabel_2}. 
In task1, we calculated the mean of all model outputs as the final ranking results.
In task2 and task3, we almost used the same network structure except the number of neurons in the classifier.
Finally, Our approach ranked 5th, 7th and 8th, respectively.
\begin{table}
  \centering \begin{tabular}{cccc}
      \toprule
      \textbf{SubTask} & \textbf{Model} & \textbf{Metric} & \textbf{Ranking} \\
      \midrule
      task1 & 6 large models & ndcg=0.9025 & 5th \\
      task2 & only 1 large model & micro f1=0.8194 & 7th \\
      task3 & only 1 large model & micro f1=0.8686 & 8th \\
      \bottomrule
  \end{tabular}
  \caption{Performance of our approach on the private leaderboard. In task1, we used six InfoXLM$_{large}$ models that fine-tuned by different datasets or methods.
  In task2 and task3, we used only one InfoXLM$_{large}$ model with the same network structure, as shown in Figure~\ref{fig:figure_2}. }
  \label{tab:tabel_2} 
\end{table}
\subsection{Ablation Studies}
We investigate the impact of adopting different pre-training task in the task2 setting.
In Table~\ref{tab:tabel_3}, we show the Mask-LM losses after 5 epochs of pre-training and Micro-F1 scores after 2 epochs of fine-tuning.
We find that the Product2Query task achieves an 0.008 improvement compared to the baseline, and the contrastive learning task doesn't get a significant gain.
\begin{table}
  \centering \begin{tabular}{lcc}
      \toprule
      \textbf{Pre-Training Task} & \textbf{CV-MLM Loss} & \textbf{CV-Micro F1} \\
      \midrule
      Mask LM & 1.966 & 74.97 \\
      +Product2Query & 1.969 & 75.05 \\
      ++Product2Brand & 1.978 & 75.08 \\
      +++Contrastive Learning & 2.047 & 75.08 \\
      \bottomrule
  \end{tabular}
  \caption{The effect of different pre-training tasks and keep accumulating from top to bottom.
   We report the cross validation MLM-Loss and Micro-F1 Score $\times$ 100 in the task2 setting.}
  \label{tab:tabel_3} 
\end{table}

As shown in Table~\ref{tab:tabel_4}, we compare several loss functions, and we adopt Poly1 loss function in task2 and task3 which based on the actual performances.
We observe that the Focal loss function and GHM loss function have worse performance than the cross-entropy loss function in the task2 setting.
\begin{table}
  \centering \begin{tabular}{lc}
      \toprule
      \textbf{Classification Loss} & \textbf{CV-Micro F1}\\
      \midrule
      CE Loss & 75.08 \\
      Focal Loss & 74.73 \\
      GHM Loss & 74.85 \\
      Poly1 Loss & 75.21 \\
      \bottomrule
  \end{tabular}
  \caption{The effect of different losses in the task2 setting.
  We report the cross validation Micro-F1 Score $\times$ 100.}
  \label{tab:tabel_4} 
\end{table}

In this subsection, we explore several methods for further improving the model's performance in fine-tuning stage.
As presented in Table~\ref{tab:tabel_5}, we adopt all of these methods to improve the model's generalization and robustness.
We observe that the exponential moving average method(EMA), adversarial training(FGM) and regularized dropout strategy(R-Drop) could improve the model's generalization and robustness.
But the Embedding Mixup strategy doesn't get a significant gain.
\begin{table}
  \centering \begin{tabular}{lc}
      \toprule
      \textbf{Methods} & \textbf{CV-Micro F1}\\
      \midrule
      +EMA & 75.19 \\
      ++FGM & 75.30 \\
      +++R-Drop & 75.43 \\
      ++++Embedding Mixup & 75.43 \\
      \bottomrule
  \end{tabular}
  \caption{The effect of different strategies and keep accumulating from top to bottom.
  We report the cross validation Micro-F1 Score $\times$ 100 in the task2 setting.}
  \label{tab:tabel_5} 
\end{table}

As shown in Table~\ref{tab:tabel_6}, we consider using smaller datasets with removing $\sim$4\% noisy labels.
We used the smaller dataset to achieve an 0.005 improvement in task1, but we get worse results in tash2 and task3.
It could be explained that since task1 contains more difficult samples, the manually annotated data contains more label errors.
\begin{table}
  \centering \begin{tabular}{lc}
      \toprule
      \textbf{Confident Learning} & \textbf{CV-Metric}\\
      \midrule
      with-in-task1 & NDCG, +0.005 \\
      with-in-task2 & Micro-F1, -0.003 \\
      with-in-task3 & Micro-F1, -0.002 \\  
      \bottomrule
  \end{tabular}
  \caption{The effect of removing ~4\% noisy labels.}
  \label{tab:tabel_6} 
\end{table}

\section{Conclusion and Future Work}
In this work, we provide an overview of the combined approach to improve the quality of search results.
We use data augmentation, multitask pretraining strategy and several fine-tuning methods to achieve considerably performance.
Specifically, we use mlm task, classification task and contrastive learning task in pre-training stage.
And we use exponential moving average method(EMA), adversarial training(FGM) and regularized dropout strategy(R-Drop) to improve the model's generalization and robustness in fine-tuning stage.
Moreover, we use a multi-granular semantic unit to discover the queries and products textual metadata for enhancing the representation of the model.
Future work of our system includes:
1) Comparing with other pre-trained language models, such as deborta$_{large}$.
2) Using other training strategies, such as self-distillation.

\bibliographystyle{ACM-Reference-Format}
\bibliography{custom-base-zhichunroad}

\appendix
\section{Appendix}
\subsection{Poisson Distribution}
\label{appendix:Poisson}
\begin{figure}[htbp]
    \centering 
    \includegraphics[width=0.47\textwidth]{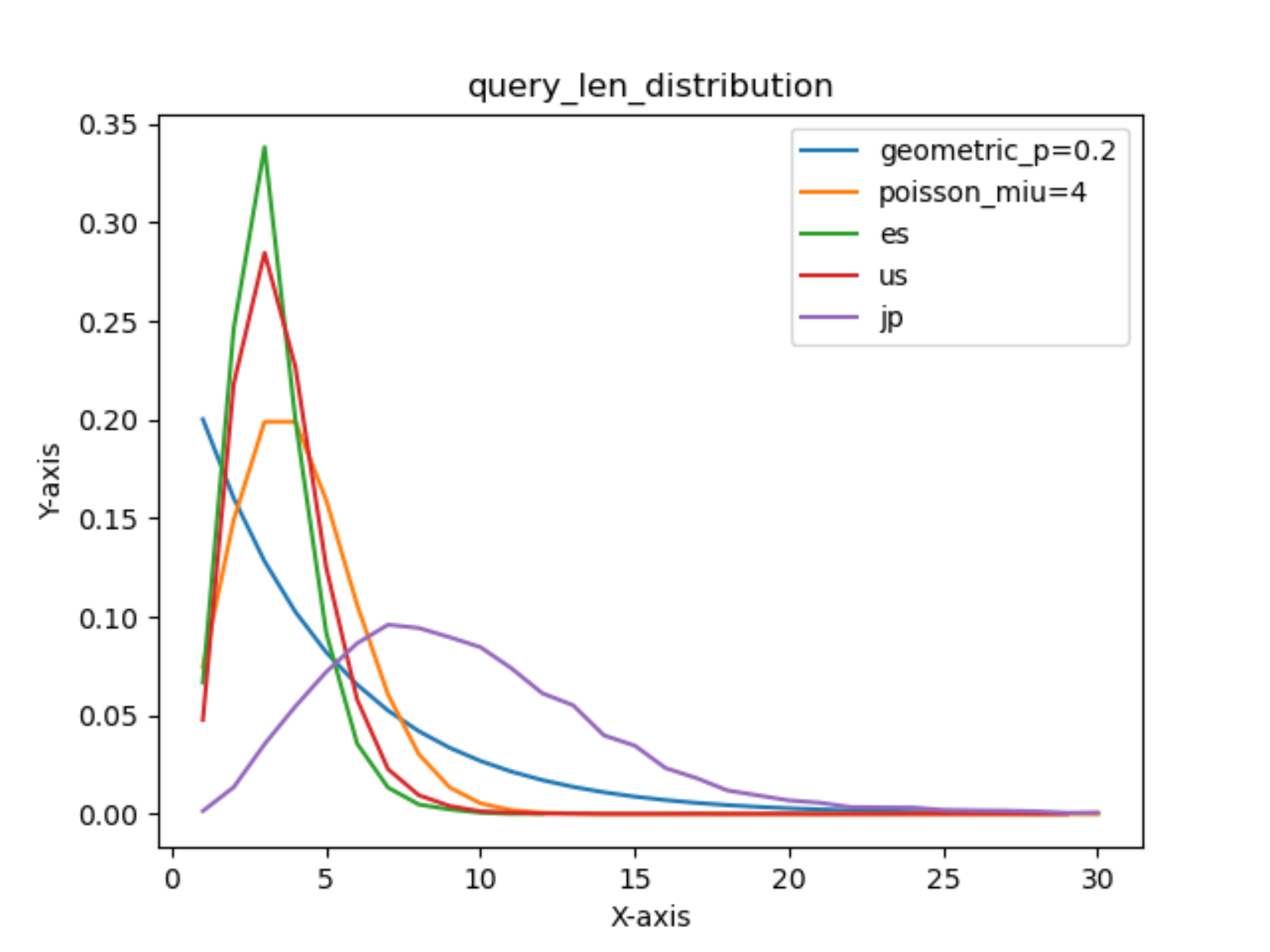}
    \caption{The length distribution of queries in different languages.}
    \label{fig:figure_3}
\end{figure}
As presented in Figure~\ref{fig:figure_3}, the median of Spanish and English queries is 14 which satisfies the Poisson distribution of $\mu$ is set to 4.
And the median of the Japanese query is 31 which satisfies the Poisson distribution with $\mu$ is set to 8.

\subsection{EmbeddingBag Initialization}
\label{appendix:embbag}
The multi-granular semantic unit implemented by Embedding-Bag\footnote{\url{https://pytorch.org/docs/stable/generated/torch.nn.EmbeddingBag.html}}. 
As presented in Table~\ref{tab:tabel_6}, the way of random initialization converges slowly, so we don’t record the final result. 
And when the Embedding-Bag is initialized by Word2vec, our approach obtain the best performance. 
\begin{table}
    \begin{tabular}{cc}
        \toprule
        \textbf{Methods} & \textbf{CV-Micro F1}\\
        \midrule
        Random$^\diamondsuit$ & - \\
        Word2vec$^\clubsuit$  & 85.33 \\
        Freeze$^\heartsuit$  & 85.29\\
        \bottomrule
    \end{tabular}
    \caption{The performance of different initialization methods of the multi-granular semantic unit. We report the cross validation Micro-F1 Score $\times$ 100 in the task3 setting.}
    \label{tab:tabel_6} 
\end{table}

\end{document}